\documentclass{article}

\usepackage{arxiv}

\renewcommand{\headeright}{}
\renewcommand{\undertitle}{}

\usepackage[utf8]{inputenc}
\usepackage[T1]{fontenc}
\usepackage{hyperref}
\usepackage{url}
\usepackage{booktabs}
\usepackage{float}
\usepackage{amsfonts}
\usepackage{amsmath}
\usepackage{amssymb}
\usepackage{nicefrac}
\usepackage{microtype}
\usepackage{lipsum}
\usepackage{graphicx}
\usepackage{xcolor}
\usepackage{pifont}
\usepackage{natbib}
\usepackage{tikz}
\usepackage{pgfplots}
\pgfplotsset{compat=1.18}
\usepackage{listings}
\usepackage[ruled,vlined]{algorithm2e}

\usetikzlibrary{arrows.meta, positioning, shapes.geometric, fit, backgrounds, calc, shadows.blur}

\graphicspath{{figures/}}

\title{eMEM: A Hybrid Spatio-Temporal Memory System for Embodied Agents}

\author{
  A.~Haroon Rasheed \quad Maria Kabtoul \\
  Automatika Robotics \\
  Inria, France
  \thanks{Correspondence: \texttt{\{haroon, maria\}@automatikarobotics.com}}
}

\date{}

\begin{document}
\maketitle

\begin{abstract}
  We present eMEM (Embodied Memory), a hybrid graph-based memory system
  for embodied agents operating in physical environments. Current
  agent memory architectures, such as Generative
  Agents~\citep{park_generative_agents_2023},
  MemGPT~\citep{packer_memgpt_2023}, and A-MEM~\citep{xu_amem_2025}, treat
  memory as text streams or knowledge graphs, but embodied agents require
  memory that is simultaneously searchable by \emph{meaning},
  \emph{space}, and \emph{time}. eMEM fills this gap with a
  multi-index architecture (\textsc{SQLite} for structured storage,
  \texttt{hnswlib} for approximate nearest-neighbour semantic search, and
  an R-tree for spatial queries) unified behind a single graph model. A tiered consolidation pipeline transforms raw perceptual
  observations into compressed summaries, mirroring hippocampal-neocortical
  consolidation in biological
  systems~\citep{mcclelland_cls_1995, kumaran_cls_updated_2016}. Ten
  agent-facing recall tools expose memory retrieval primitives, including
  concept-to-location resolution and cross-layer recall, as first-class
  operations for LLM tool calling. The system is fully embedded and runs in-process alongside the agent.
  In addition we introduce \emph{eMEM-Bench v1}, a benchmark we construct
  over ProcTHOR-10K~\citep{deitke_procthor_2022} scenes for embodied memeory evaluation. The benchmark is organised explicitly around eight cognitive-psychology paradigms (DRM lures,
  pattern separation, pattern completion, source monitoring,
  context-dependent retrieval, long-horizon interference, serial
  position, and a foil-augmented retention curve), each chosen so that
  the result is interpretable against the broader memory-systems
  literature in humans and prior agent-memory systems; a level of
  diagnostic that surface-task benchmarks like LoCoMo or OpenEQA cannot
  provide. eMEM scores 80.8 weighted-mean over 988 probes, and ablations
  trace this to the memory architecture rather than the language model:
  it leads a pure RAG baseline by 19\,points, and a
  Generative-Agents-style memory
  stream~\citep{park_generative_agents_2023} by 16--20\,points on the
  binding and discrimination paradigms most central to embodiment ---
  source monitoring, pattern separation, and pattern completion. We
  release both the system and the benchmark
  code.\footnote{Code: \url{https://github.com/automatikarobotics/emem}}
\end{abstract}

\keywords{embodied memory \and spatio-temporal memory \and retrieval-augmented generation \and robotics \and consolidation \and LLM tool use}

\section{Introduction}
\label{sec:introduction}

Embodied agents (i.e. robots, drones, autonomous vehicles) accumulate
experience that is inherently spatial and temporal. A robot
patrolling a building does not just need to remember \emph{what} it saw;
it needs to remember \emph{where} and \emph{when} it saw it, relate
observations across perceptual modalities, and compress weeks of patrol
data into retrievable knowledge.

Current LLM-agent memory systems, while increasingly sophisticated, are
designed around text-based conversational agents. Generative
Agents~\citep{park_generative_agents_2023} maintains a timestamped memory
stream with reflections, MemGPT~\citep{packer_memgpt_2023} pages
information between context and archival storage, and
A-MEM~\citep{xu_amem_2025} organises memories as interconnected
Zettelkasten notes; Mem0~\citep{chhikara_mem0_2025} and
Zep~\citep{rasmussen_zep_2025} extend this line with production-oriented
long-term stores and temporal knowledge graphs. None of these systems
natively handles spatial coordinates, multi-modal perception layers, or
the kind of spatial reasoning required by embodied agents (``where is
the kitchen?'', ``what objects are near the charging station?'').
Robotics research has in parallel produced sophisticated spatial
representations, e.g. 3D scene graphs such as
Hydra~\citep{hughes_hydra_2022} and
ConceptGraphs~\citep{gu_conceptgraphs_2024}, visual-language
maps~\citep{huang_vlmaps_2023}, and semantic
mapping~\citep{raychaudhuri_semantic_mapping_2025}; these are, however,
typically \emph{perceptual} systems focused on the current state of the
environment, not \emph{memory} systems that track how the environment
changes over time, consolidate experience into knowledge, or support
episodic recall. eMEM bridges this gap by combining the temporal and
consolidation mechanisms of LLM-agent memory with the spatial grounding
of robotic scene understanding.

Each of eMEM's components is, individually, established art: SQLite,
HNSW~\citep{malkov_hnsw_2020}, and R-tree indices are standard;
DBSCAN~\citep{ester_dbscan_1996} is an algorithm from 1996;
LLM-summarisation-as-consolidation is implemented by Generative Agents / MemGPT /
A-MEM; and ReAct-over-tools was presented by \citet{yao_react_2023}.
The novelty lies in their combination: spatial grounding, multi-layer
perception, and tiered consolidation brought together in a single
embedded package, and an embodied-memory implementation that performs
across the range of paradigms identified in biological memory research.
We state this explicitly so readers can calibrate the contribution
against prior work.

eMEM is shapaed by four design commitments. First, spatial grounding is
not optional. Every observation, entity, and summary carries a position,
and spatial context is a first-class retrieval axis alongside semantic
similarity and temporal order. Second, memories are organized in layers.
An embodied agent may run multiple perception models simultaneously
(e.g. a VLM describing scenes, an object detector listing entities, a
place classifier labelling regions), and eMEM stores these as
independent, co-located observations on different layers rather than
forcing them into a hierarchy, such that the structure between the layers is emergant and not enforced. Third, places emerge from data. There is
no dedicated ``region'' or ``place'' primitive. If a VLM consistently
describes coordinates around $(10, 10)$ as ``kitchen'', the kitchen
exists as a cluster of observations and gists, and queries like ``where
is the kitchen?'' are resolved by semantic search and spatial
aggregation over the existing primitives. Fourth, we think in recall
patterns. When a person remembers ``the kitchen'', they recall where it
is (spatial), what it is like (cross-modal synthesis), and what
happened there recently (temporal); new capabilities should therefore
emerge from new query shapes over existing primitives, not from new
primitives.

Our contributions are as follows.
\begin{itemize}
  \item We introduce a hybrid graph-based memory architecture with four
  node types and six edge types, unified over three indices (SQLite,
  HNSW, R-tree) so that semantic, spatial, and temporal queries resolve
  against the same underlying store (\S\ref{sec:architecture}).
  \item We develop a tiered consolidation pipeline
  (working $\to$ short-term $\to$ long-term $\to$ archived) with
  two-phase archival and DBSCAN-clustered time-window summarisation,
  giving each observation a biologically motivated lifecycle
  (\S\ref{sec:consolidation}).
  \item We expose a 10-tool retrieval surface covering
  concept-to-location resolution, spatial-semantic fusion, and
  cross-layer recall as first-class LLM tool calls
  (\S\ref{sec:tool-interface}).
  \item We introduce \emph{eMEM-Bench v1}, an embodied-memory
  benchmark built over ProcTHOR-10K
  scenes~\citep{deitke_procthor_2022} and organised explicitly
  around eight cognitive-psychology paradigms drawn from human-memory
  research --- DRM
  lures~\citep{roediger_mcdermott_drm_1995}, pattern
  separation~\citep{yassa_stark_pattern_sep_2011} and completion,
  source monitoring~\citep{johnson_source_monitoring_1993},
  context-dependent retrieval~\citep{godden_context_1975},
  long-horizon interference, serial
  position~\citep{murdock_serial_1962}, and a foil-augmented
  retention curve~\citep{ebbinghaus_1885}. Each paradigm is chosen so
  that it isolates a distinct memory function with an established
  empirical signature, has a natural embodied-agent analogue, and is
  programmatically generable at scale. Existing agent-memory
  benchmarks are either text-only (LoCoMo, conversational dialogue)
  or category-defined (OpenEQA, perception-grounded QA); neither
  exercises the embodied axes (spatial, multi-modal, interoceptive)
  while being interpretable against the broader memory-systems
  literature (\S\ref{sec:evaluation}).
  \item As a methodological contribution we introduce a
  \emph{foil-augmented} probe construction for embodied-agent
  benchmarks: yes/no probes paired with cross-scene absent foils,
  which expose otherwise-hidden bias and instruction-parroting
  failure modes (\S\ref{sec:foils}, \S\ref{sec:limitations}).
\end{itemize}

The rest of the paper is organised as follows.
\S\ref{sec:architecture} describes the architecture, data model, and
indices. \S\ref{sec:consolidation} details the consolidation pipeline.
\S\ref{sec:tool-interface} presents the LLM tool surface.
\S\ref{sec:neuroscience} relates eMEM's design to neuroscience evidence
on memory and spatial representation. \S\ref{sec:related-work} positions
eMEM against prior work. \S\ref{sec:evaluation} introduces eMEM-Bench
v1, justifies the paradigm panel, and reports experimental
results including the cross-system ablation against pure RAG.
\S\ref{sec:limitations} discusses limitations and future
directions.

\section{Architecture}
\label{sec:architecture}

\begin{figure}[t]
  \centering
  \begin{tikzpicture}[
    node distance = 6mm and 8mm,
    facade/.style = {draw, rounded corners, thick, minimum width=52mm, minimum height=8mm, fill=blue!8, align=center, font=\small},
    mid/.style    = {draw, rounded corners, minimum width=30mm, minimum height=12mm, fill=gray!8, align=center, font=\small},
    store/.style  = {draw, rounded corners, minimum width=78mm, minimum height=8mm, fill=green!8, align=center, font=\small},
    idx/.style    = {draw, minimum width=22mm, minimum height=8mm, fill=white, align=center, font=\footnotesize},
    arr/.style    = {-Latex, thick}
  ]
    \node[facade] (facade) {\textsc{SpatioTemporalMemory} facade};

    \node[mid, below=of facade, xshift=-32mm] (wm)   {WorkingMemory\\\footnotesize in-process buffer};
    \node[mid, below=of facade]               (tools){Tools\\\footnotesize 10 retrieval primitives};
    \node[mid, below=of facade, xshift=32mm]  (cons) {Consolidation\\\footnotesize gist + entity};

    \node[store, below=15mm of tools] (store) {\textsc{MemoryStore} (hybrid persistence)};
    \node[idx, below=2mm of store.south, xshift=-26mm] (sql)  {SQLite\\structured};
    \node[idx, below=2mm of store.south]               (hnsw) {HNSW\\semantic};
    \node[idx, below=2mm of store.south, xshift=26mm]  (rt)   {R-tree\\spatial};

    \draw[arr] (facade) -- (wm);
    \draw[arr] (facade) -- (tools);
    \draw[arr] (facade) -- (cons);
    \draw[arr] (wm)    -- (store);
    \draw[arr] (tools) -- (store);
    \draw[arr] (cons)  -- (store);
  \end{tikzpicture}
  \caption{eMEM system overview. The \textsc{SpatioTemporalMemory} facade
  orchestrates a working-memory buffer, a retrieval tool surface, and a
  consolidation engine, all backed by a hybrid \textsc{MemoryStore}
  combining SQLite, HNSW, and R-tree indices.}
  \label{fig:architecture}
\end{figure}
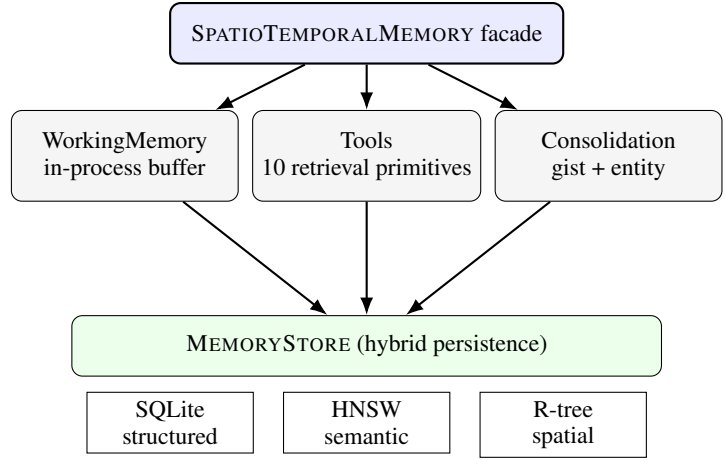

Figure~\ref{fig:architecture} shows the overall system layout. A
\textsc{SpatioTemporalMemory} system orchestrates three cooperating
subsystems (a working-memory buffer, a consolidation engine, and a
10-tool retrieval surface) on top of a hybrid \textsc{MemoryStore} that
combines SQLite, HNSW, and R-tree indices. All subsystems run
in-process, and the only persistence state
is a SQLite file together with a binary HNSW index on disk.

\begin{figure}[t]
  \centering
  \begin{tikzpicture}[
    obs/.style    = {draw, circle, thick, minimum size=13mm, fill=blue!10, align=center, font=\footnotesize},
    epi/.style    = {draw, rectangle, rounded corners, thick, minimum width=18mm, minimum height=9mm, fill=orange!15, align=center, font=\footnotesize},
    gist/.style   = {draw, diamond, thick, aspect=1.6, inner sep=1pt, fill=green!15, align=center, font=\footnotesize},
    ent/.style    = {draw, rectangle, thick, minimum width=18mm, minimum height=9mm, fill=red!10, align=center, font=\footnotesize},
    edg/.style    = {-Latex, thick, font=\scriptsize},
    co/.style     = {Latex-Latex, thick, dashed, font=\scriptsize}
  ]
    \node[epi] (e1) at (2.2, 3.0){Episode$_a$};
    \node[epi] (e2) at (7.0, 3.0){Episode$_b$};

    \node[obs] (o1) at (0.6, 0.5)  {Obs$_1$};
    \node[obs] (o2) at (3.6, 0.5)  {Obs$_2$};

    \node[ent] (en1) at (8.0, 0.5) {Entity$_1$};
    \node[ent] (en2) at (12.5, 0.5) {Entity$_2$};

    \node[gist](g1) at (2.2, -2.0){Gist};

    \draw[edg] (o1) -- node[pos=0.55, left=2pt]{BELONGS\_TO} (e1);
    \draw[edg] (o2) -- (e1);

    \draw[edg] (e1) -- node[above]{FOLLOWS} (e2);

    \draw[edg] (e2) to[bend left=40] node[above]{SUBTASK\_OF} (e1);

    \draw[edg] (g1) -- node[pos=0.5, left=2pt]{SUMMARIZES} (o1);
    \draw[edg] (g1) -- (o2);

    \draw[edg] (en1) -- node[above]{OBSERVED\_IN} (o2);

    \draw[co]  (en1) -- node[above]{COOCCURS\_WITH} (en2);
  \end{tikzpicture}
  \caption{Data model. Four node types (observation, episode, gist,
  entity) connected by six edge types. Gists summarise groups of
  observations; entities are tracked across observations and linked by
  co-occurrence.}
  \label{fig:data-model}
\end{figure}
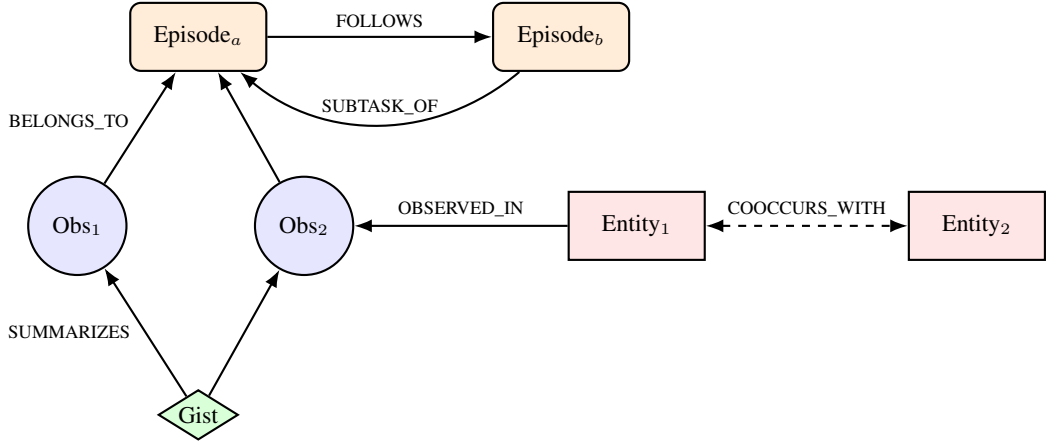

\subsection{Data Model}
\label{sec:data-model}

eMEM represents memory as a graph with four node types and six edge
types (Figure~\ref{fig:data-model}). Observation nodes carry a single
perceptual event recorded with text, coordinates, timestamp, and a layer
tag; they are the analogue of a hippocampal episodic trace. Episode
nodes are task containers that group observations and support
hierarchical nesting, playing the role of episodic context. Gist nodes
are consolidated summaries with spatial extent (a centre and a radius)
produced from a cluster of observations, and play the role of a
neocortical schema. Entity nodes are persistent object concepts tracked
across observations and merged under a learned similarity threshold,
giving the system a semantic-memory store of objects. These nodes are
related by six edge types: \texttt{BELONGS\_TO} links an observation to
its episode; \texttt{FOLLOWS} and \texttt{SUBTASK\_OF} relate episodes
to one another; \texttt{SUMMARIZES} links a gist to the observations it
subsumes; \texttt{OBSERVED\_IN} records the observation in which an
entity was seen; and \texttt{COOCCURS\_WITH} records co-presence between
entities.

\subsection{Memory Tiers}
\label{sec:memory-tiers}

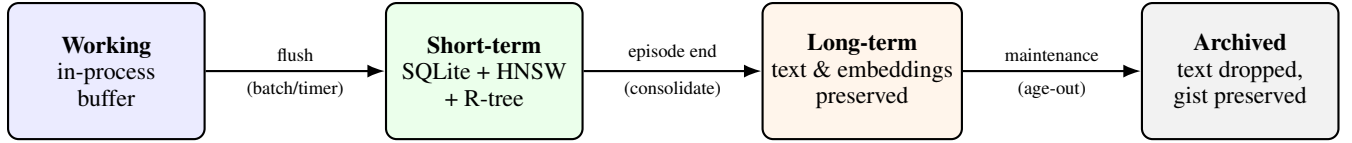
\begin{figure}[t]
  \centering
  \begin{tikzpicture}[
    tier/.style = {draw, rounded corners, thick, minimum width=26mm, minimum height=18mm, align=center, font=\small},
    arr/.style  = {-Latex, thick}
  ]
    \node[tier, fill=blue!8]   (w)   at (0, 0)     {\textbf{Working}\\\footnotesize in-process\\\footnotesize buffer};
    \node[tier, fill=green!8]  (s)   at (5.0, 0)   {\textbf{Short-term}\\\footnotesize SQLite + HNSW\\\footnotesize + R-tree};
    \node[tier, fill=orange!8] (l)   at (10.0, 0)  {\textbf{Long-term}\\\footnotesize text \& embeddings\\\footnotesize preserved};
    \node[tier, fill=gray!10]  (a)   at (15.0, 0)  {\textbf{Archived}\\\footnotesize text dropped,\\\footnotesize gist preserved};

    \draw[arr] (w) -- node[above,font=\scriptsize]{flush} node[below,font=\scriptsize]{(batch/timer)} (s);
    \draw[arr] (s) -- node[above,font=\scriptsize]{episode end} node[below,font=\scriptsize]{(consolidate)} (l);
    \draw[arr] (l) -- node[above,font=\scriptsize]{maintenance} node[below,font=\scriptsize]{(age-out)}    (a);
  \end{tikzpicture}
  \caption{Memory tiers. Observations are encoded into a working buffer,
  flushed to short-term storage where they are searchable, consolidated
  into long-term memory at episode boundaries (with text and embeddings
  preserved), and eventually archived (i.e. leaving only a gist
  searchable) once they age out.}
  \label{fig:memory-tiers}
\end{figure}

Memory in eMEM progresses through four tiers, paralleling the biological
consolidation pathway from hippocampus to
neocortex~\citep{mcclelland_cls_1995, diekelmann_sleep_memory_2010}; see
Figure~\ref{fig:memory-tiers}. At the working tier, observations are
held in an in-process deque buffer that auto-flushes on batch size
(default 5) or time interval (default 2\,s), with entity extraction
triggered periodically from the flush callback. On flush, observations
enter the short-term tier, where they are stored across SQLite, HNSW,
and the R-tree and become fully searchable while awaiting consolidation.
At episode end they are promoted to the long-term tier with text and
embeddings preserved; raw observations therefore remain searchable
alongside their gists, which ensures that no detail is lost during
active operation. After a configurable age a \texttt{maintenance()}
sweep moves them to the archived tier, dropping raw text and embeddings
and leaving only the gist searchable, which bounds long-term storage
growth; a necessity for physical agents expected to operate continuously over an arbitrarily long period.

\subsection{Three-Index Storage}
\label{sec:storage}

eMEM engages three complementary indices, each tuned to a different
retrieval axis, unified under a single store interface. SQLite holds
structured data: observation text, timestamps, layer names, tier status,
episode membership, entity attributes, and graph edges. It is indexed
on timestamp, layer, episode, tier, and entity name, and provides the
relational backbone for filtering and joins. HNSW~\citep{malkov_hnsw_2020} (via
\texttt{hnswlib}) provides approximate nearest-neighbour semantic search
over embedding vectors; a single index stores embeddings for
observations, gists, and entities, with an ID mapping in SQLite tracking
node type. We use cosine distance, build parameters $M{=}16$ and
$\mathrm{ef_{construction}}{=}200$, and search $\mathrm{ef}{=}50$ by
default, and the binary format persists across process restarts. The
R-tree (via \texttt{rtree}) supports 3D spatial range and
nearest-neighbour queries; it is rebuilt on startup from SQLite
coordinates, and string UUIDs are mapped to integer IDs internally.
The rebuild cost is $O(N \log N)$ in the number of stored
observations. A typical query such as \textit{``what did the robot see near the kitchen recently?''} engages all three indices: HNSW finds semantically
relevant memories, the R-tree filters by spatial proximity, and SQLite
applies layer and time constraints.

Semantic search can optionally be biased towards recent observations.
When the recency weight $\alpha > 0$, the score combines HNSW distance
with temporal recency as
$\mathrm{score} = d_{\mathrm{HNSW}} + \alpha \cdot t_{\mathrm{age}} /
t_{\mathrm{halflife}}$,
which is useful for queries such as ``where is the \texttt{<name\_of\_an\_object>}?'' that should
prefer the latest sighting. The weight is zero by default so that the
standard ranking is purely semantic.

\subsection{Multi-Layer Perception}
\label{sec:multi-layer}

Observations carry a \texttt{layer\_name} tag that identifies their
perceptual source. Layers are user-defined and represent a semantic context i.e. different questions a perception stack asks about the environment; the same
coordinates may therefore carry observations on several layers
simultaneously. A VLM layer might record ``a kitchen with white cabinets
and a wooden table'', a detections layer ``chair, table, refrigerator,
microwave'', a place identification layer ``kitchen'', and a human-presence layer ``empty room'', all at the same position. These are four independent
observations rather than attributes of a single entry: all query tools
accept an optional layer filter, the \texttt{get\_current\_context} tool
groups nearby observations by layer for structured presentation, and
the consolidation engine uses layer-aware synthesis for multi-layer
clusters.

\section{Consolidation}
\label{sec:consolidation}

eMEM's consolidation engine transforms raw observations into compressed
long-term knowledge, implementing the core insight of Complementary
Learning Systems theory~\citep{mcclelland_cls_1995,
kumaran_cls_updated_2016}: rapid encoding of specific episodes
(observations) followed by gradual extraction of regularities (gists).

\subsection{Episode Consolidation}
\label{sec:episode-consolidation}

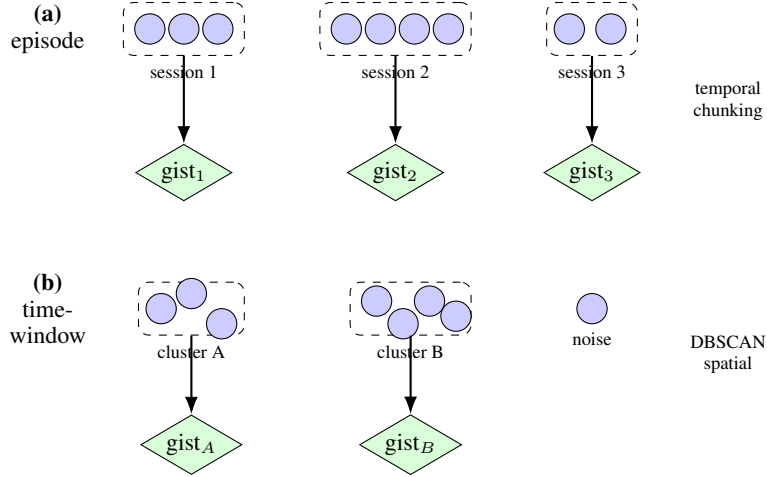
\begin{figure}[t]
  \centering
  \begin{tikzpicture}[
    obs/.style   = {draw, circle, minimum size=4mm, inner sep=0pt, fill=blue!20},
    gist/.style  = {draw, diamond, aspect=1.7, inner sep=1pt, fill=green!15, font=\footnotesize},
    grp/.style   = {draw, rounded corners, dashed, inner sep=4pt, minimum height=7mm},
    lbl/.style   = {font=\footnotesize, align=center},
    arr/.style   = {-Latex, thick},
    tag/.style   = {font=\scriptsize, align=center}
  ]
    \node[lbl] at (-1.4, 1.0) {\textbf{(a)}\\episode};

    \node[grp, minimum width=16mm] (c1) at (0.4, 1.0)  {};
    \node[obs] at (-0.05, 1.0) {}; \node[obs] at (0.4, 1.0) {}; \node[obs] at (0.85, 1.0) {};
    \node[tag, below=0mm of c1] {session 1};

    \node[grp, minimum width=20mm] (c2) at (3.2, 1.0) {};
    \node[obs] at (2.55, 1.0){}; \node[obs] at (3.00, 1.0){}; \node[obs] at (3.45, 1.0){}; \node[obs] at (3.90, 1.0){};
    \node[tag, below=0mm of c2] {session 2};

    \node[grp, minimum width=12mm] (c3) at (5.8, 1.0) {};
    \node[obs] at (5.5, 1.0) {}; \node[obs] at (6.05, 1.0) {};
    \node[tag, below=0mm of c3] {session 3};

    \node[gist] (g1) at (0.4, -0.9) {gist$_1$};
    \node[gist] (g2) at (3.2, -0.9) {gist$_2$};
    \node[gist] (g3) at (5.8, -0.9) {gist$_3$};
    \draw[arr] (c1) -- (g1);
    \draw[arr] (c2) -- (g2);
    \draw[arr] (c3) -- (g3);
    \node[tag] at (7.6, 0.05) {temporal\\chunking};

    \node[lbl] at (-1.4, -2.7) {\textbf{(b)}\\time-\\window};

    \node[grp, minimum width=14mm] (ca) at (0.5, -2.7) {};
    \node[obs] at (0.1, -2.7){}; \node[obs] at (0.5, -2.5){}; \node[obs] at (0.9, -2.9){};
    \node[tag, below=0mm of ca] {cluster A};

    \node[grp, minimum width=16mm] (cb) at (3.4, -2.7) {};
    \node[obs] at (2.95, -2.6){}; \node[obs] at (3.30, -2.9){}; \node[obs] at (3.65, -2.6){}; \node[obs] at (4.00, -2.8){};
    \node[tag, below=0mm of cb] {cluster B};

    \node[obs] at (5.8, -2.7) {};
    \node[tag] at (5.8, -3.1) {noise};

    \node[gist] (gA) at (0.5, -4.5) {gist$_A$};
    \node[gist] (gB) at (3.4, -4.5) {gist$_B$};
    \draw[arr] (ca) -- (gA);
    \draw[arr] (cb) -- (gB);
    \node[tag] at (7.6, -3.3) {DBSCAN\\spatial};
  \end{tikzpicture}
  \caption{Consolidation. (a) At episode end, observations are temporally
  chunked (default gap: 30\,min) into sessions; each chunk produces a
  focused gist. (b) For non-episodic observations that age past the
  consolidation window, DBSCAN spatial clustering produces one gist per
  cluster; ungrouped observations remain as-is until archived.}
  \label{fig:consolidation}
\end{figure}

When an episode ends, its observations pass through four stages, the
first three illustrated in Figure~\ref{fig:consolidation}a. Observations
are first sorted by timestamp and split into temporal chunks wherever
the gap between consecutive observations exceeds a configurable window
(default 30\,min); this ensures that a long episode, e.g. a
long running task spanning hours, produces several focused
gists rather than one monolithic summary. Each chunk is then summarised
into a separate gist via an LLM, with a concatenation fallback
when no LLM is configured, and multi-layer chunks are presented to the
summariser grouped by layer so that it can produce a structured
cross-layer synthesis. Alongside summarisation, entities are extracted
from every episode observation and either merged with existing entities
(by semantic similarity and spatial proximity) or created afresh.
Finally, the observations themselves are promoted to the long-term tier
with text and embeddings preserved, so that they remain searchable
alongside their gists.

\subsection{Two-Phase Archival}
\label{sec:archival}

Consolidation follows a two-phase gradient inspired by biological
consolidation during sleep~\citep{diekelmann_sleep_memory_2010,
born_consolidation_sleep_2012}. At episode end the first phase
promotes observations to the long-term tier with text, embeddings, and
spatial entries all preserved, so that both the raw observation and the
gist remain searchable; the raw provides detail while the gist provides
context. The second phase runs later as a maintenance sweep over
observations older than \texttt{archive\_after\_seconds} (default one
hour, though a deployment might set this to days): text and embeddings
are dropped and only the gist remains searchable. An internal
\texttt{maintenance()} method triggers this phase, which in a
deployed system would run during idle periods, loosely analogous to
sleep-dependent consolidation in biological memory.

\subsection{Time-Window Consolidation}
\label{sec:time-window}

Not all observations arrive inside an episode. For those that age past
the consolidation window outside any episode, a DBSCAN
clustering~\citep{ester_dbscan_1996} groups
observations by spatial proximity and each cluster is consolidated
independently (Figure~\ref{fig:consolidation}b). This avoids merging
observations from distinct locations into incoherent summaries, and it
naturally produces one gist per place visited during a free-running
interval.

\subsection{Entity Tracking}
\label{sec:entity-tracking}

Entities are persistent object representations that survive
consolidation. Extraction happens at two points. At flush time,
observations accumulate in an entity buffer; after a configurable number
of flushes (default 10) or a configurable time interval (default 60\,s),
whichever comes first, the buffer is drained and entities are
extracted.
Entities are also extracted during episode consolidation itself,
guarded by the deduplication flag. When the LLM identifies an object,
the engine checks for existing entities with similar names (cosine
similarity above 0.85 by default) at nearby positions (within 5\,m by default); matches are
merged, with coordinates averaged, observation counts incremented, and
temporal bounds extended, and new entities are created otherwise.
Co-occurrence is tracked via \texttt{COOCCURS\_WITH} edges, enabling
queries such as \textit{``what objects appear together with the red
chair?''}.

\section{Systems Characterisation}
\label{sec:systems}

Running memory in-process and for arbitrarily long periods of time motivates measuring concrete numbers on
ingest throughput, storage footprint, retrieval fidelity, and startup
cost. We measure on an NVIDIA RTX A5000 workstation using a synthetic
random-unit-vector embedder (no embedding-model cost) at three
observation counts: $N = 10^3$, $10^4$, $10^5$. The vector
dimension is fixed at 384, matching the default
\texttt{embedding\_dim}. All measurements use the default
configuration (Table~\ref{tab:config-defaults}) except where noted.

\begin{table}[t]
  \centering
  \small
  \setlength{\tabcolsep}{4pt}
  \renewcommand{\arraystretch}{1.1}
  \begin{tabular}{r r r r r r r r}
    \toprule
    & \multicolumn{1}{c}{\textbf{Ingest}} & \multicolumn{3}{c}{\textbf{HNSW recall@10}} & \multicolumn{1}{c}{\textbf{R-tree}} & \multicolumn{2}{c}{\textbf{Size}} \\
    \cmidrule(lr){3-5} \cmidrule(lr){7-8}
    $N$ & (obs/s) & $\mathrm{ef}{=}16$ & $\mathrm{ef}{=}50$ & $\mathrm{ef}{=}200$ & rebuild (s) & SQLite (MiB) & HNSW (MiB) \\
    \midrule
    $10^3$  & 2\,558 & 0.59 & 0.89 & 1.00 & 0.02 & 0.72   & 1.61   \\
    $10^4$  & 947    & 0.15 & 0.34 & 0.75 & 0.42 & 6.27   & 16.07  \\
    $10^5$  & 561    & 0.02 & 0.06 & 0.21 & 5.25 & 62.34  & 160.64 \\
    \bottomrule
  \end{tabular}
  \caption{Systems benchmarks. Ingest throughput excludes
  LLM-inference cost (entity extraction disabled; synthetic
  random-unit-vector embeddings). HNSW recall@10 measured against a
  brute-force cosine top-10 over 500 random queries. R-tree rebuild
  measured by reopening a closed database. SQLite and HNSW file
  sizes are post-ingestion on-disk footprints.}
  \label{tab:systems-benchmark}
\end{table}

Three observations deserve emphasis. First, storage grows
predictably: the SQLite file and HNSW index together consume
$\approx 23$\,MiB at $N = 10^4$ and $\approx 223$\,MiB at $N = 10^5$
on a 384-dimensional store. Linear extrapolation gives a rough upper
bound of $\approx 2.2$\,GiB at $10^6$ --- manageable on a robot's
onboard storage, but enough to motivate the archival phase of the
consolidation pipeline (\S\ref{sec:archival}). Second, R-tree rebuild
at startup is sub-second at $10^4$ and $\approx 5$\,s at $10^5$,
consistent with the $O(N \log N)$ growth
(\S\ref{sec:storage}); an incremental persistence path is a
straightforward extension for much longer-lived deployments. Third,
ingest throughput declines with $N$ (from $\sim 2.6$\,k obs/s at
$10^3$ to $\sim 0.6$\,k obs/s at $10^5$), which tracks the HNSW
insertion cost rather than any SQLite or R-tree ceiling.

These numbers exclude entity extraction (which would add an
LLM-inference cost per flush interval) and other LLM-side consolidation
cost. The separable LLM-inference cost is reported alongside the
benchmark results in \S\ref{sec:evaluation}.

\section{LLM Tool Interface}
\label{sec:tool-interface}

eMEM exposes memory retrieval to an agent as a set of tools; each is a
callable with a JSON-schema description suitable for LLM tool
use~\citep{yao_react_2023}. Tools return formatted strings optimised for
token efficiency.

\begin{table}[t]
  \centering
  \small
  \renewcommand{\arraystretch}{1.15}
  \begin{tabular}{l l p{78mm}}
    \toprule
    \textbf{Tool} & \textbf{Axis} & \textbf{Role} \\
    \midrule
    \texttt{semantic\_search}    & meaning     & Unified ANN search across observations, gists, and entities. \\
    \texttt{spatial\_query}      & location    & Observations within a radius of a 3D point. \\
    \texttt{temporal\_query}     & time        & Observations in a time range, chronologically ordered. \\
    \texttt{episode\_summary}    & episodes    & Gist summaries of a completed episode. \\
    \texttt{search\_gists}       & long-term   & Semantic search restricted to consolidated gists. \\
    \texttt{entity\_query}       & objects     & Query persistent entity records (by name or area). \\
    \texttt{get\_current\_context}& situation  & Nearby observations grouped by layer, plus gists, entities, and recent activity. \\
    \texttt{body\_status}        & body        & Latest interoception readings (battery, temperature, joints). \\
    \midrule
    \texttt{locate}              & concept$\to$place & Resolve a semantic concept to a centroid and spread. \\
    \texttt{recall}              & cross-modal       & Chain \texttt{locate} $+$ spatial $+$ gists $+$ entities for a full recall. \\
    \bottomrule
  \end{tabular}
  \caption{eMEM's 10-tool retrieval surface. The top block lists the eight
  core primitives covering meaning, location, time, and body state. The
  bottom block lists two higher-order tools that compose the primitives.}
  \label{tab:tools}
\end{table}

The core tools (top block of Table~\ref{tab:tools}) cover the three
primary retrieval axes. \texttt{semantic\_search}, \texttt{search\_gists},
and \texttt{entity\_query} address the meaning axis;
\texttt{spatial\_query} addresses location; and \texttt{temporal\_query}
and \texttt{episode\_summary} address time. Two further primitives
address situational awareness (\texttt{get\_current\_context}, which
returns nearby observations grouped by layer together with gists,
entities, and recent activity) and body state (\texttt{body\_status},
which returns the latest interoception readings).

Two higher-order tools sit on top of these primitives and implement
recall patterns that would otherwise require multi-step reasoning from
the agent. \texttt{locate(concept)} resolves a semantic concept to a
spatial position by running \texttt{semantic\_search}, collecting
coordinates from matching observations, gists, and entities, and
returning a centroid, a spread radius, and a match count; it therefore
bridges \textit{``kitchen''} (a concept) and $(10.3, 10.1, 0.0)$ (a
position). \texttt{recall(query)} implements \textit{``tell me about
X''} by chaining \texttt{locate} with spatial aggregation: it locates
the concept, runs \texttt{spatial\_query} at that location across every
layer, gathers consolidated gists by area, and pulls tracked entities,
formatting the result grouped by layer to produce a cross-modal summary
grounded in spatial reality. This composition pattern, treating
retrieval primitives as building blocks from which higher-order recall
emerges, is how we apply the design commitment to think in recall
patterns: new capabilities come from new query shapes, not new
primitives.

\section{Neuroscience Foundations}
\label{sec:neuroscience}

eMEM is not a model of the brain, but many of its design choices
are informed by computational counterparts of well-established
findings in the neuroscience of memory; we sketch that mapping here.

The Complementary Learning Systems
theory~\citep{mcclelland_cls_1995, kumaran_cls_updated_2016} posits two
cooperating systems: the hippocampus rapidly encodes specific episodes
while the neocortex gradually extracts statistical regularities through
consolidation. eMEM's observation nodes play the role of the
hippocampal traces (specific, timestamped, spatially grounded) and gist
nodes play the role of the neocortical abstractions. The four-tier
pipeline (working $\to$ short-term $\to$ long-term $\to$ archived)
implements the CLS consolidation gradient more granularly than the
two-system biological model, but preserves its essential dynamic: rich
episodic traces are progressively compressed into general knowledge,
with a two-phase archival that matches the observation that recently
consolidated memories remain accessible for a period before full
compression~\citep{diekelmann_sleep_memory_2010,
born_consolidation_sleep_2012}. A recent RAG-oriented system,
HippoRAG~\citep{gutierrez_hipporag_2024}, also draws on the CLS framing
but applies it to text-document retrieval via personalised PageRank
rather than to embodied spatio-temporal memory.

A related line of work concerns the hippocampal spatial coordinate
system for memory, from O'Keefe's place
cells~\citep{okeefe_place_cells_1971} and the grid cells of the
entorhinal cortex~\citep{hafting_grid_cells_2005} to a recent
perspective~\citep{benna_fusi_place_cells_2021} that reframes place
cells as general-purpose memory cells whose spatial tuning \emph{emerges}
from compressing sequential experience. Our commitment that places
emerge from data follows precisely this view: the kitchen is not a
dedicated data structure but a cluster of observations that share
spatial proximity and semantic content, surfaced at query time by
\texttt{locate}. Tulving's distinction between episodic memory (specific
events in spatio-temporal context) and semantic memory (general
knowledge abstracted from episodes)~\citep{tulving_episodic_1972,
tulving_elements_1983} maps directly onto eMEM's node taxonomy:
observations and episodes carry the \emph{when} and \emph{where} of
particular events, while gists and entities hold the knowledge
abstracted from them, and the consolidation pipeline is the
computational route from one to the other.

The space-from-sequence intuition has a recent algorithmic instantiation in clone-structured cognitive graphs~\citep{george_cscg_2021}, generalised to a theory of the hippocampus as a latent-sequence learner~\citep{raju_latent_sequence_2024}. The same family of place-cell and splitter-cell behaviour is recovered from a single mechanism: a cloned hidden Markov model trained on egocentric, action-conditioned observations. This position further holds that Euclidean coordinates are not merely emergent but, an artefact, with the latent graph alone supporting recall, planning, and transfer. eMEM, on the other hand, uses coordinates because a robot already has them, an engineering choice rather than a philosophical one. Both perspectives agree on graph-shaped memory and on consolidation as the bridge from episodic to semantic, and the latent-graph machinery composes naturally with our explicit spatial axis. We return to this convergence in \S\ref{sec:limitations}.

Two further findings motivate the way retrieval and update work in
eMEM. Retrieved memories become labile and must be reconsolidated,
allowing new information to integrate into existing
traces~\citep{nader_reconsolidation_2000}; the entity auto-merge step
implements exactly this, so a re-observed entity updates its record
(coordinates, observation count, temporal bounds) rather than producing
a duplicate. Likewise, retrieval is enhanced when context matches
encoding~\citep{godden_context_1975,
tulving_encoding_specificity_1973}, and \texttt{get\_current\_context}
exploits this by using the agent's current position as an automatic
retrieval cue, mirroring how returning to a place tends to trigger
recall of experiences from there.

Finally, interoception is increasingly recognised as integral to memory
rather than peripheral to it. Interoceptive predictive
coding~\citep{seth_interoception_2013, seth_critchley_2013} argues that
the brain maintains predictive models of internal body states and that
these states are not metadata but integral to encoding, and the
hippocampus itself integrates interoceptive signals alongside spatial
and temporal information~\citep{quigley_interoception_2021}. Following
this view, and building on the INTERO
framework~\citep{maimon_intero_2025} and the ArmarX memory
system~\citep{pellerkonrad_armarx_2023}, eMEM treats body state
(battery, CPU temperature, joint health) as peer observation layers
that flow through the same \texttt{ObservationNode $\to$ GistNode}
pipeline as world observations. Co-location with the agent's pose means
that \texttt{spatial\_query} near a steep ramp naturally surfaces both
``steep ramp'' (VLM) and ``rapid battery drain'' (interoception), so
that spatial-interoceptive associations emerge from the data without
explicit edges.

We stress that these correspondences are motivational rather than
literal. eMEM is an engineering system, and the neuroscience motivation serves to
explain why its interface shape is a reasonable one for an embodied
agent, not to make claims about the brain.

\section{Related Work}
\label{sec:related-work}

eMEM sits at the intersection of three research threads: LLM-agent
memory, embodied spatial representations, and biological memory models.
Table~\ref{tab:related-work} summarises the capability axes on which
eMEM differs from prominent prior work.

\begin{table}[t]
  \centering
  \small
  \renewcommand{\arraystretch}{1.1}
  \begin{tabular}{l c c c c c}
    \toprule
    \textbf{System} & \textbf{Spatial} & \textbf{Temporal} & \textbf{Consol.} & \textbf{Multi-layer} & \textbf{Embedded} \\
    \midrule
    \multicolumn{6}{l}{\textit{LLM-agent memory}} \\
    Generative Agents~\citep{park_generative_agents_2023} & \ding{55} & \checkmark & \checkmark & \ding{55} & \ding{55} \\
    MemGPT~\citep{packer_memgpt_2023}                     & \ding{55} & \checkmark & \ding{55} & \ding{55} & \checkmark \\
    A-MEM~\citep{xu_amem_2025}                            & \ding{55} & \checkmark & \checkmark & \ding{55} & \ding{55} \\
    Mem0~\citep{chhikara_mem0_2025}                       & \ding{55} & \checkmark & \checkmark & \ding{55} & \ding{55} \\
    Zep~\citep{rasmussen_zep_2025}                        & \ding{55} & \checkmark & \checkmark & \ding{55} & \ding{55} \\
    \midrule
    \multicolumn{6}{l}{\textit{Embodied spatial representations}} \\
    Hydra~\citep{hughes_hydra_2022}                       & \checkmark & $\ast$ & \ding{55} & \checkmark & \checkmark \\
    ConceptGraphs~\citep{gu_conceptgraphs_2024}           & \checkmark & $\ast$ & \ding{55} & \checkmark & \checkmark \\
    VLMaps~\citep{huang_vlmaps_2023}                      & \checkmark & \ding{55} & \ding{55} & \ding{55} & \checkmark \\
    SayPlan~\citep{rana_sayplan_2023}                     & \checkmark & \ding{55} & \ding{55} & \ding{55} & \ding{55} \\
    \midrule
    \textbf{eMEM (ours)}                                  & \checkmark & \checkmark & \checkmark & \checkmark & \checkmark \\
    \bottomrule
  \end{tabular}
  \caption{Capability comparison. \textit{Spatial}: supports 3D coordinate
  queries. \textit{Temporal}: observation timestamps as a first-class
  query axis. \textit{Consol.}: background extraction of higher-level
  summaries. \textit{Multi-layer}: multiple perception modalities
  co-located. \textit{Embedded}: runs in-process, no external server.
  ($\ast$) Hydra and ConceptGraphs maintain persistent object
  instances with update timestamps and therefore capture some temporal
  dynamics at the scene-graph level; they do not, however, expose
  temporal \emph{query} primitives over an experience stream, which is
  the contrast we draw here.}
  \label{tab:related-work}
\end{table}

\subsection{LLM-agent memory}

Generative Agents~\citep{park_generative_agents_2023} introduced the
memory stream with reflections, and eMEM adopts the same core shape
(observations plus consolidation into higher-level summaries); it adds
spatial coordinates, multi-layer perception, and entity tracking, and
it provides recency, importance, and relevance as explicit query axes
rather than a single weighted score. MemGPT~\citep{packer_memgpt_2023}
introduced tiered memory with self-directed paging, and eMEM's four
tiers extend this with automatic consolidation and spatial awareness.
MemoryBank~\citep{zhong_memorybank_2024} applies the Ebbinghaus
forgetting curve, while eMEM implements an analogous decay through tier
promotion and archival. A-MEM~\citep{xu_amem_2025} organises memories
as interconnected Zettelkasten notes, and eMEM's graph achieves similar
connectivity but with spatial and temporal grounding. The
production-focused systems Mem0~\citep{chhikara_mem0_2025} and
Zep~\citep{rasmussen_zep_2025} share the graph-based philosophy; eMEM is a production grade framework for robotics and differentiates from these systems with its spatial index, multi-layer memory model, and embodied focus.
HippoRAG~\citep{gutierrez_hipporag_2024} is an especially relevant recent
addition that makes the hippocampal-CLS framing explicit in an LLM-RAG
setting via personalised PageRank over an open knowledge graph; eMEM
shares the CLS-inspired motivation and the graph-based retrieval surface
but targets embodied memory with spatial indexing and multi-layer
perception rather than text-document retrieval. The memory survey
of~\citet{zhang_memory_survey_2024} provides a taxonomy onto which
eMEM's architecture maps cleanly.

\subsection{Embodied spatial representations}

Hydra~\citep{hughes_hydra_2022} builds hierarchical 3D scene graphs in
real time from sensor data. eMEM does not replicate this geometric
scene graph; it instead operates at a higher level, storing \emph{what
the agent remembers} about places rather than the current geometric
state. The two systems are therefore complementary, with Hydra providing
perception and eMEM providing memory.
ConceptGraphs~\citep{gu_conceptgraphs_2024} constructs open-vocabulary
3D scene graphs using foundation models; eMEM's entity nodes with
VLM-generated descriptions are analogous but add temporal tracking
(\texttt{first\_seen}, \texttt{last\_seen}, observation counts) and
consolidation. SayPlan~\citep{rana_sayplan_2023} showed that LLMs can
plan over graph-structured spatial representations, and eMEM's tool
interface provides exactly that query substrate.
EmbodiedRAG~\citep{booker_embodiedrag_2024} applies retrieval-augmented
generation to 3D scene graphs by surfacing task-relevant subgraphs, and
eMEM's \texttt{get\_current\_context} and \texttt{recall} tools
implement a similar selective-retrieval pattern.
VLMaps~\citep{huang_vlmaps_2023} fuses CLIP embeddings with spatial
reconstructions at the voxel level; eMEM uses the same idea at the
episodic level, enabling temporal reasoning that voxel-grid
representations do not support.

\subsection{Embodied QA benchmarks}

Several benchmarks evaluate aspects of embodied question answering
without isolating the memory system.
OpenEQA~\citep{majumdar_openeqa_2024} presents over 1{,}600
human-written questions on real-world scenes, but it sends video frames
directly to a VLM and therefore bypasses any structured memory.
SQA3D~\citep{ma_sqa3d_2023} tests situated spatial reasoning from a
fixed pose, not memory recall over time.
FindingDory~\citep{yadav_findingdory_2025} does isolate memory from
exploration, but it uses task-oriented navigation scoring on VLMs
directly with image sequences.
LMEE-Bench~\citep{wang_explore_longterm_2026} couples exploration
efficiency with memory recall accuracy. eMEM-Bench complements these
benchmarks by replaying fixed trajectories into a structured memory
system and testing six distinct dimensions, including interoception and
entity tracking, that none of the above cover.

Most LLM-agent memory systems lack spatial grounding, and most robotic
spatial representations lack temporal memory, interoception and consolidation. eMEM
combines both, and we ground this combination in the neuroscience
framing discussed in \S\ref{sec:neuroscience}.

\section{Evaluation: eMEM-Bench v1}
\label{sec:evaluation}

In order to evaluate eMEM effectively, we created \emph{eMEM-Bench v1}, a benchmark constructed over
ProcTHOR-10K~\citep{deitke_procthor_2022} scenes that is organised
explicitly around eight cognitive-psychology paradigms drawn from the
human-memory literature.

\subsection{Why a paradigm based embodied benchmark}
\label{sec:why-paradigm}

Existing agent-memory benchmarks fall into two groups, each with a
mismatch to the embodied-agent setting. LoCoMo~\citep{maharana_locomo_2024}
and similar long-horizon dialogue benchmarks evaluate text-only
conversational memory and do not exercise spatial grounding,
multi-modal perception, or interoception; the axes that motivate
embodied memory in the first place. Embodied-QA benchmarks such as
OpenEQA~\citep{majumdar_openeqa_2024} test perception-grounded
question answering in scenes, but the questions are surface-task
defined (``what colour is the chair?'') and do not isolate which
\emph{memory function} produces a correct or incorrect answer.

A category-defined benchmark (``temporal'', ``spatial'',
``cross-layer'') tells us \emph{where} a system fails. A
\emph{paradigm}-defined benchmark tells us \emph{which cognitive
function} is implicated, and makes the result interpretable against
a century of published findings in human memory. eMEM-Bench v1
operationalises eight paradigms, each tied to a specific empirical
phenomenon and a specific class of failure mode that matters for
embodied agents over long deployments. The paradigms were chosen on
three criteria: \emph{(i)} each isolates a distinct memory function
with a well-known empirical signature; \emph{(ii)} each has a natural
embodied-agent analogue, i.e. it is not a contrived translation of a
human task; and \emph{(iii)} each can be operationalised
programmatically over ProcTHOR scenes without per-question human
authoring, so that the benchmark scales.

\subsection{Harness}
\label{sec:benchmark-harness}

eMEM-Bench v1 uses a replay-based harness. It ingests pre-recorded
trajectories into a fresh eMEM instance, drives a
ReAct~\citep{yao_react_2023} agent against eMEM's tool surface to
answer paradigm probes, and scores predictions against ground truth
with an LLM judge. The ReAct loop is capped at five steps, and a
post-processor strips leaked meta-commentary and maps unanswerable
responses to empty strings. Experiments use Qwen 3.6~(27B)~\citep{qwen3_2025}
as both agent and entity-extractor,
\texttt{gemma3:27b} as the judge (the judge is held constant across
all configurations), and
\texttt{nomic-embed-text-v2-moe}~\citep{nussbaum_nomic_embed_v2_2025}
embeddings. No proprietary APIs are in the loop.

\subsection{Scenes and trajectories}
\label{sec:scenes}

v1 uses 20 multi-room ProcTHOR-10K
houses~\citep{deitke_procthor_2022} captured in
AI2-THOR~\citep{kolve_ai2thor_2017}. Each scene is toured along a
fixed teleport-and-rotate path that visits every reachable position
and rotates $360^\circ$ at each waypoint. Three perception layers are
recorded at each view. VLM scene description, object detections,
and place classification, alongside synthetic interoception
(battery, CPU temperature). Trajectories are recorded once and
replayed deterministically; different memory configurations differ
only in what their memory system surfaces from the \emph{same}
observation stream. This isolates memory effects from perception
variance.

Each paradigm has a fixed \emph{schedule}: a sequence of
\texttt{IngestPhase} / \texttt{AdvanceClockPhase} / \texttt{ProbePhase}
operations that control what the agent ingests, what virtual delay
elapses, and what is asked. \texttt{AdvanceClockPhase} can fire
consolidation and archival passes; this lets a paradigm test recall
\emph{after} the memory has been compressed by hours, days, weeks, or
years of simulated time.

\subsection{The eight paradigms}
\label{sec:paradigms}

The paradigms group into three families. We describe each below with
its source in human memory research, the empirical signature it
tests, and the embodied-agent failure mode it surfaces.

Every probe in the v1 release was reviewed manually by the authors
after programmatic generation; candidates with ambiguous ground
truth, perception-unfriendly items, or inconsistent scene metadata
were discarded before the harness was run.

\paragraph{Encoding and similarity.}

\textit{DRM} (Deese-Roediger-McDermott; \citealt{roediger_mcdermott_drm_1995}).
Subjects study a list of associated words (\emph{bed, rest, awake,
tired, dream, \ldots}) and reliably ``remember'' a never-studied lure
(\emph{sleep}). The signature is high false-recognition of
semantically related lures. \emph{Embodied analogue:} when a VLM
describes a kitchen with \emph{counter, sink, faucet, fridge,
kettle}, an agent without lure-rejection machinery is liable to
report a \emph{microwave} in that kitchen. We instantiate the
paradigm by selecting four objects unique to one room of a scene,
mixing them with a semantically similar lure absent from the room,
and asking the agent to verify each.

\textit{Pattern separation} (\citealt{yassa_stark_pattern_sep_2011};
hippocampal CA3/DG).
The signature is the ability to encode similar episodes as distinct
traces. \emph{Embodied analogue:} two different houses visited
consecutively have overlapping object lists; the agent must say
``the wine bottle was in House~B, not House~A'' rather than
collapsing them. We pair similar ProcTHOR houses and probe
attribution of objects to one house or the other.

\textit{Pattern completion} (the hippocampal complement to pattern
separation). The signature is recovering a complete memory from a
partial cue. \emph{Embodied analogue:} an operator says ``the mug''
and the agent must recover ``the blue mug on the kitchen counter at
14:32''. Almost every real-world retrieval uses partial cues; this
paradigm tests whether the agent can resolve them to specific
encoded items.

\paragraph{Source and context.}

\textit{Source monitoring}~(\citealt{johnson_source_monitoring_1993}).
The signature is the ability to attribute a memory to its source ---
externally perceived vs.\ internally generated, modality~A vs.\ B,
which speaker said it. \emph{Embodied analogue:} ``Did you learn
about the bread from the VLM scene description or from the object
detector?''. This is the embodied analogue of reality monitoring;
failure means an agent cannot tell its detector hallucinations from
its perceptions, with safety consequences.

\textit{Context-dependent retrieval}~(\citealt{godden_context_1975}).
The Godden-Baddeley signature is that recall is best in the context
of encoding (divers recall items learned underwater better when
tested underwater). \emph{Embodied analogue:} ``Is there a chair
here?'' must condition on the agent's \emph{current} position, not
return any chair from anywhere in the scene. This is the most
common retrieval pattern for an operating embodied agent, and an
explicit test that the system surfaces spatial co-location as a
first-class retrieval axis.

\paragraph{Temporal and interference.}

\textit{Long-horizon interference.} Classical memory research
distinguishes \emph{proactive} interference (earlier material
hurts later) from \emph{retroactive} (later material hurts earlier).
\emph{Embodied analogue:} two house tours separated by a simulated
day-long gap; the agent must answer questions about House~Alpha
after ingesting House~Beta. This is the canonical
multi-day-deployment test: does yesterday's patrol survive today's
overlay?

\textit{Serial position}~(\citealt{murdock_serial_1962}). The
signature is the U-shaped recall function over a sequence: primacy
(early items recalled well) and recency (late items recalled well),
with a depressed middle. \emph{Embodied analogue:} a chain of
multiple house tours; the agent is asked when in the chain a given
object first appeared (early/middle/late). This tests whether the
memory preserves temporal ordering across many ingested episodes.

\textit{Retention curve}~(\citealt{ebbinghaus_1885}). The classical
forgetting curve: recall as a function of delay since encoding. We
probe at six delays $\{1\text{h}, 1\text{d}, 1\text{wk}, 1\text{mo},
6\text{mo}, 1\text{yr}\}$ of simulated time,
firing consolidation/archival at each step
so that the probe at $+1\text{yr}$ sees memory state after a year of
the pipeline's pressure, not after raw ingestion. This paradigm
characterises a long-running deployment.

\subsection{Foil methodology}
\label{sec:foils}

The retention curve uses 50/50 yes/no probes; foils are drawn from
a \emph{cross-scene absent pool}; objects that appear (and were
confirmed unique to the room) in some other scene in the benchmark, but are absent from
this scene's ProcTHOR ground-truth inventory. The pool guarantees
foil items are concrete, perception-friendly, and unambiguously
absent. Without foils, a yes/no recall paradigm collapses into a
test of yes-bias. Balancing the polarity
restores signal detection discriminability (hit rate vs.\
false-alarm rate per delay bucket)
and converts the paradigm into a test of recall.

We propose foil-augmentation as a general methodological discipline
for embodied-agent benchmarks: any yes/no probe should be balanced
against an absent-foil pool, and any free-text probe whose ground
truth is fully recoverable from a single stored observation should
be re-engineered so the agent must verify perception, not just
retrieve text.

\subsection{Results}
\label{sec:results}

\begin{table}[t]
  \centering
  \small
  \begin{tabular}{l l r r r r}
    \toprule
    \textbf{Paradigm} & \textbf{Family} & $N$ & \textbf{Score} & \textbf{Perfect\,\%} & \textbf{Tool\,Acc} \\
    \midrule
    DRM~\citep{roediger_mcdermott_drm_1995}                    & Encoding / similarity     &  60 & 86.7 & 86.7 & 0.67 \\
    Pattern separation~\citep{yassa_stark_pattern_sep_2011}    & Encoding / similarity     &  53 & 71.7 & 71.7 & 0.52 \\
    Pattern completion                                          & Encoding / similarity     &  73 & 79.5 & 75.3 & 0.43 \\
    Source monitoring~\citep{johnson_source_monitoring_1993}    & Source / context          &  70 & 78.6 & 71.4 & 0.53 \\
    Context-dep.\ retrieval~\citep{godden_context_1975}& Source / context          & 136 & 63.8 & 61.8 & 0.30 \\
    Long-horizon interference                                   & Temporal / interference   &  62 & 69.4 & 69.4 & 0.48 \\
    Serial position~\citep{murdock_serial_1962}                 & Temporal / interference   &  54 & 69.4 & 66.7 & 0.46 \\
    Retention curve (foils)~\citep{ebbinghaus_1885}             & Temporal / interference   & 480 & 89.1 & 87.1 & 0.68 \\
    \midrule
    \textbf{Overall (weighted)}                                 &                           & \textbf{988} & \textbf{80.8} & 78.5 & 0.56 \\
    \bottomrule
  \end{tabular}
  \caption{eMEM-Bench v1 results across the eight cognitive paradigms
  retained for evaluation, over 20 multi-room ProcTHOR-10K
  scenes~\citep{deitke_procthor_2022}. Scored 1--5 by an LLM judge
  (gemma3:27b), normalised to 0--100. ``Perfect\,\%'' is the share of
  probes scored 5. ``Tool\,Acc'' is the fraction of probes for which
  the agent invoked at least one tool the paradigm marks as expected.}
  \label{tab:emem-bench-v1}
\end{table}

Table~\ref{tab:emem-bench-v1} reports the eight paradigms over
988~probes, with an overall weighted mean of 80.8. Encoding and similarity paradigms cluster
between 71 and 87, with DRM at the top: the consolidation pipeline
produces gists that suppress associated-lure activation, the
cognitive signature the paradigm was designed to detect. Source
monitoring scores 78.6, with eMEM's per-layer tagging making the
candidate layers directly addressable at retrieval time. This
embodied analogue of source attribution succeeds for the same
reason it does in humans, namely that the relevant distinction is
encoded at storage rather than reconstructed at query. Pattern
separation and pattern completion sit at 72 and 79, indicating that
the explicit observation/episode/gist taxonomy gives the agent the
right material to discriminate similar scenes and to complete
partial cues. Context-dependent retrieval scores a lower 63.8; here the
limiting factor is the agent's tool selection under a spatial
constraint rather than the memory representation. This provides a
tractable agent-side improvement step. Temporal and interference paradigms
range from 69 to 89, topped by the retention curve.

The retention curve is the clearest demonstration of long-term
stability: hit rate holds at ceiling on seen items from 1\,h to
1\,yr of simulated delay, so eMEM shows no
measurable forgetting on room-unique items across a year of virtual
time (as intended by the design),  semantic clustering and spatial co-location keep
salient items stable under consolidation and archival pressure. False
alarms stay near $15.5\%$ rather than rising with delay, and are
item-level: a handful of near-synonym confusions from entity
extraction (``key chain'' versus ``keys''), not forgetting.

\subsection{Ablations}
\label{sec:ablation}

The ablations in Table~\ref{tab:ablation} ask two questions: how much
the memory architecture adds over a simpler memory the same language
model could use, and which part of the architecture is responsible.
Every condition runs at a matched cap of six schedules per paradigm,
so each number is a within-paradigm comparison against the full system
on the same probes. Two conditions swap in an external baseline.
Flat-RAG gives the agent semantic search over a vector store and a
timestamp filter and nothing more, the configuration an agent-memory
architect reaches for by default. Gen-Agents adds a recency-weighted
memory stream and a current-context query, after
\citet{park_generative_agents_2023}. The third condition, no-hybrid,
leaves the full system and all ten tools in place and changes nothing
the agent can see; it only switches off the lexical half of retrieval,
so ranking falls back to dense vectors alone.

\begin{table}[t]
  \centering
  \small
  \setlength{\tabcolsep}{4.5pt}
  \begin{tabular}{l r r r r r r r r}
    \toprule
    & & \multicolumn{1}{c}{\textbf{Full}}
      & \multicolumn{2}{c}{\textbf{Flat-RAG}}
      & \multicolumn{2}{c}{\textbf{Gen-Agents}}
      & \multicolumn{2}{c}{\textbf{No-hybrid}} \\
    \cmidrule(lr){3-3}\cmidrule(lr){4-5}\cmidrule(lr){6-7}\cmidrule(lr){8-9}
    \textbf{Paradigm} & $N$ & Score & Score & $\Delta$ & Score & $\Delta$ & Score & $\Delta$ \\
    \midrule
    Serial position          & 13 & 69.2 & 69.2 & $+0.0$  & 84.6 & $\mathbf{+15.4}$ & 15.4 & $\mathbf{-53.9}$ \\
    Source monitoring        & 22 & 83.0 & 61.4 & $-21.6$ & 62.5 & $-20.5$ & 47.7 & $\mathbf{-35.2}$ \\
    Context-dep.\ retrieval  &  6 & 83.3 & 33.3 & $\mathbf{-50.0}$ & 83.3 & $+0.0$  & 83.3 & $+0.0$  \\
    Pattern completion       & 30 & 90.0 & 66.7 & $-23.3$ & 71.7 & $-18.3$ & 85.0 & $-5.0$  \\
    Pattern separation       & 53 & 67.9 & 60.9 & $-7.1$  & 51.4 & $-16.5$ & 66.5 & $-1.4$  \\
    DRM (false memory)       & 19 & 84.2 & 57.9 & $-26.3$ & 73.7 & $-10.5$ & 89.5 & $+5.3$  \\
    Long-horizon interf.     & 54 & 70.4 & 57.4 & $-13.0$ & 70.4 & $+0.0$  & 64.8 & $-5.6$  \\
    Retention decay          & 36 & 97.2 & 86.1 & $-11.1$ & 83.3 & $-13.9$ & 97.2 & $+0.0$  \\
    \midrule
    \textbf{Mean}            &    & \textbf{80.7} & \textbf{61.6} & $\mathbf{-19.0}$
                                  & \textbf{72.6} & $\mathbf{-8.0}$
                                  & \textbf{68.7} & $\mathbf{-12.0}$ \\
    \bottomrule
  \end{tabular}
  \caption{Ablations on eMEM-Bench v1, scored by the same LLM judge as
  Table~\ref{tab:emem-bench-v1}. Each cell runs at a matched cap of six
  schedules per paradigm, so every entry is a within-paradigm
  comparison to the \emph{Full} column; $N$ is the resulting probe
  count. \emph{Flat-RAG} restricts the agent to semantic search and a
  timestamp filter. \emph{Gen-Agents} is a recency-weighted memory
  stream after \citet{park_generative_agents_2023}. \emph{No-hybrid}
  keeps the full system and all ten tools but disables the lexical half
  of retrieval, leaving dense-vector ranking. $\Delta$ is the change
  from \emph{Full}; the mean is the unweighted average over the eight
  paradigms.}
  \label{tab:ablation}
\end{table}

Flat-RAG is the weaker of the two baselines, nineteen points behind the
full system. It fails where plain similarity search has nothing to fall
back on: it cannot tie retrieval to the agent's current location, so
context-dependent retrieval drops fifty points, and it has only
embedding distance with which to reject a tempting lure, so DRM drops
twenty-six. Given just a vector store and a clock, the same language
model is a much weaker agent.

Gen-Agents is the stronger baseline, and the more telling one. Its
current-context query restores context-dependent retrieval, and its
append-only log even edges past the full system on serial position,
where recovering the order of events is just reading positions off the
log. But that is the one thing a chronological store gets for free, and
it is also the paradigm that matters least once an agent has to do more
than recite a timeline. It flatters the average score which trails by eight points. On the paradigms that carry the weight of
embodied memory --- separating similar scenes, completing a partial
cue, attributing a memory to its source --- it falls sixteen to twenty
points short, because a stream ranked by recency and similarity cannot
tell two close memories apart or fix one to its origin. The same line
runs through both baselines: retrieving by meaning and reading off
recency are cheap, while discriminating and binding memories are what
the structured index in eMEM enables.

No-hybrid places that contribution in a single mechanism. Its
twelve-point average is concentrated rather than spread: serial
position drops fifty-four points and source monitoring thirty-five,
while the semantic paradigms hold steady or improve. Serial position
and source monitoring both ask the agent to retrieve one particular
item: the object seen at a specific point in the tour, or the
observation that came from a specific source. Dense-vector search is
poorly suited to this, because it ranks memories by overall similarity
of meaning, so many near-identical observations score almost the same
and the correct one is not reliably on top. The lexical half of
retrieval breaks these ties by matching specific words, and that is
what recovers the right item. The semantic paradigms never rely on it:
rejecting a DRM lure only requires finding the right neighbourhood of
meaning rather than one exact memory, which dense similarity already
does, so removing the lexical half leaves them unchanged.

\paragraph{Reading the numbers.}
The paradigm framing makes the 80.8 weighted mean directly
interpretable against published findings in human memory and in
prior agent-memory systems. On each paradigm eMEM lands on the
productive side of the diagnostic the paradigm was designed to
expose. Where the score is lower, the paradigm itself
points to a concrete next step. Context-dependent retrieval, for
instance, implicates the agent's spatial-tool selection more than
the memory representation, i.e not a structural gap. This
level of diagnostic resolution is what surface-task benchmarks
cannot deliver, and is the reason for organising the evaluation
around paradigms in the first place. It is pertinent to remember that this evaluation has been performed with open source small models that can run on a consumer GPU. Both query formulation and tool selection are significantly impacted by model quality, making it non-trivial to measure performance of the memory system in islation.

\section{Limitations and Future Work}
\label{sec:limitations}

\paragraph{Agent-side improvements.}
A meaningful fraction of the headroom on eMEM-Bench~v1 lives in the
agent rather than in the memory primitive. As noted earlier in the
\S\ref{sec:evaluation} section, the evaluation uses open-source models small
enough to fit on a consumer GPU; query formulation, tool selection,
and multi-step composition all scale with model quality, so
improvements there should lift scores across the board without
changing the memory architecture. Context-dependent retrieval (63.8)
is the clearest case: the limit is the agent's ability to select
spatial tools under a spatial constraint, not what the memory
exposes. The most common low-score pattern more generally is
semantic-search phrasing failing to match an observation's embedding
(\textit{``when did I last see the stool?''} fails when the VLM
described the object as ``bar seating''), with the symmetric
near-synonym confusions (``key chain'' vs.\ ``keys'') visible in the
retention-curve false alarms. Scaling the agent, fine-tuning a small
model specifically for memory tool use, embeddings tuned for
embodied language, and query expansion are all orthogonal to the eMEM
architecture and would compose with it.

\paragraph{Richer consolidation.}
The current pipeline weighs observations uniformly within a time
window, so an agent that spent 20 minutes in a kitchen and 30
seconds glimpsing a bathroom produces a gist that mentions only the
kitchen. Biologically, consolidation is modulated by novelty and
reward~\citep{lee_jung_consolidation_rl_2024}; adaptive weighting
that favours transitions and surprise would help briefly visited
areas. Relatedly, the \texttt{recall} tool returns observations grouped by
perception layer rather than fused into a single description per
location; a dedicated cross-layer synthesis step would lift
descriptive and episodic queries, at the cost of an additional
inference call per location.

\paragraph{Object-frame coordinates.}
A separable improvement is to bring object-frame coordinates into
the spatial axis. eMEM records observation positions; entity
tracking provides centroid positions over multiple sightings, but
single-observation queries still report the robot's pose. Depth
estimation or object-relative positioning, integrated through the
perception stack rather than the memory, would close this gap and
directly raise context-dependent retrieval. This sits at the
boundary of memory and perception; the natural composition is for
the perception stack to publish object-frame coordinates as another
layer that eMEM ingests.

\paragraph{Latent representations and schema transfer.}
eMEM is symbolic: it stores what happened and what is known, as
post-perception linguistic descriptions, not raw sensory data, latent
representations, or procedural knowledge. A direction worth pursuing
is the latent-graph tradition in computational
neuroscience~\citep{george_cscg_2021, raju_latent_sequence_2024}, in
which a cloned hidden Markov model trained on action-conditioned
sequences recovers phenomena similar to place and splitter-cells, from
latent inference alone, and supports \emph{schema transfer} by
freezing the latent transition graph as a content-free topology and
rebinding only the observation slots in a new environment.

The most promising direction for future work will be to stack the two
layers such that the observable spatio-temporal graph continues to serve as
the language-facing surface that the LLM queries; a latent-state
layer underneath assigns each observation to a clone. This should allow automatic
disambiguation of perceptually similar contexts (two visually
similar kitchens in different houses becoming distinct latent states
without manual scene tagging) and schema-based generalisation
across environments. Concretely, a robot deployed to a new house
could carry over what it has learned about kitchen topology and
rebind only the visual identities, rather than starting from scratch
each time. In this way a long-running embodied agent should accumulate transferable structure across the
environments it visits. From the current results, this extension should improve performance on pattern separation, context-dependent retrieval and the retention-curve false alarms.

\bibliographystyle{plainnat}
\bibliography{references}

\clearpage
\appendix
\section{Additional Limitations and Future Directions}
\label{sec:appendix-limitations}

Temporal reasoning is the weakest category on both benchmarks. The
agent sees timestamps in tool output but struggles to use them for
ordering, duration estimation, or first/last classification. Dedicated
temporal tools (e.g. a timeline view that renders the sequence of
visits to a location, or explicit \texttt{first\_seen} and
\texttt{last\_seen} primitives) could remove the need for the agent to
perform temporal reasoning itself.

eMEM does not support shared memory across multiple agents. Multi-agent
spatial memory with conflict resolution and perspective-taking is an
open challenge for collaborative robotics and is a natural direction
for future work. Closer to perception, eMEM stores point coordinates
rather than meshes, voxels, or occupancy grids; integration with
geometric systems such as Hydra~\citep{hughes_hydra_2022} or
ConceptGraphs~\citep{gu_conceptgraphs_2024} would provide richer
spatial reasoning without changing the memory architecture, since the
geometric representations would simply become another perception layer.

A timing consideration for entity extraction: the two entity-extraction triggers
(\texttt{entity\_extract\_flush\_interval}, default 10, and
\texttt{entity\_extract\_time\_interval}, default 60\,s) interact to
bound the latency between an observation being added and the
corresponding entity becoming available via \texttt{entity\_query}.
With default \texttt{flush\_batch\_size}~=~5 and a steady observation
rate $r$, the flush-interval trigger dominates at high $r$ (entities
are extracted every $5 \times 10 = 50$ observations) and the time
trigger dominates at low $r$ (every 60\,s regardless of buffer
fill). Worst-case availability latency for a newly observed entity is
therefore bounded below by $\min(50/r, 60\,\mathrm{s})$. Deployments
needing lower-latency entity availability can reduce both intervals
at the cost of more frequent LLM calls.

Finally, eMEM-Bench v1 evaluates on 20 multi-room ProcTHOR-10K
houses with programmatically generated, manually reviewed
questions. Extending to real-world robot trajectories, more diverse
environments, and naturalistic human-authored questions would
improve ecological validity.

\section{Implementation Details}
\label{sec:appendix-implementation}

Storage in eMEM uses only embedded libraries: SQLite (stdlib),
\texttt{hnswlib} for approximate nearest-neighbour search,
\texttt{rtree} for 3D spatial queries, \texttt{scikit-learn} DBSCAN for
time-window consolidation, and NumPy for vector operations. Embeddings
are produced through a pluggable \texttt{EmbeddingProvider} protocol
rather than a fixed model: any backend that maps text to vectors can be
supplied. The experiments in this paper use
\texttt{nomic-embed-text-v2-moe}~\citep{nussbaum_nomic_embed_v2_2025}
(768-dimensional) served locally over Ollama; an in-process
\texttt{sentence-transformers} backend
(SBERT~\citep{reimers_sbert_2019}) is also provided for deployments
that prefer to avoid a separate embedding service. A complete eMEM
instance persists as two files: a SQLite database and an HNSW binary
index. The code is organised into a small number of modules:
\texttt{types.py} (node and edge types, tiers), \texttt{config.py}
(tunable parameters), \texttt{embeddings.py} (the
\texttt{EmbeddingProvider} protocol and implementations),
\texttt{spatial.py} (the R-tree wrapper), \texttt{working\_memory.py}
(the in-process buffer with auto-flush and entity callbacks),
\texttt{store.py} (hybrid persistence over SQLite, HNSW, and the
R-tree), \texttt{consolidation.py} (temporal chunking, gist generation,
entity extraction, and archival), \texttt{tools.py} (the 10 LLM-facing
tools), and \texttt{memory.py} (the \texttt{SpatioTemporalMemory}
facade). The default configuration parameters used throughout the
paper are listed in Table~\ref{tab:config-defaults}.

\begin{table}[H]
  \centering
  \small
  \begin{tabular}{l l l}
    \toprule
    \textbf{Parameter} & \textbf{Default} & \textbf{Meaning} \\
    \midrule
    \texttt{embedding\_dim}                  & 384         & Embedding vector dimension (auto-set by the provider) \\
    \texttt{working\_memory\_size}           & 50          & Max buffered observations \\
    \texttt{flush\_batch\_size}              & 5           & Batch-size auto-flush threshold \\
    \texttt{flush\_interval}                 & 2.0\,s      & Time-based auto-flush threshold \\
    \texttt{consolidation\_window}           & 1800\,s     & Temporal gap threshold (episode chunking \& time-window) \\
    \texttt{consolidation\_spatial\_eps}     & 3.0\,m      & DBSCAN clustering radius \\
    \texttt{consolidation\_min\_samples}     & 3           & Min.\ cluster size for spatial consolidation \\
    \texttt{archive\_after\_seconds}         & 3600\,s     & Time in long-term before archival \\
    \texttt{entity\_extract\_flush\_interval}& 10          & Extract entities every $N$ flushes \\
    \texttt{entity\_extract\_time\_interval} & 60.0\,s     & Extract entities every $N$ seconds \\
    \texttt{entity\_similarity\_threshold}   & 0.85        & Cosine similarity for entity merge \\
    \texttt{entity\_spatial\_radius}         & 5.0\,m      & Spatial proximity for entity merge \\
    \texttt{recency\_weight}                 & 0.0         & Recency-weighting $\alpha$ (0 disables) \\
    \texttt{recency\_halflife}               & 3600\,s     & Time constant for recency decay \\
    \bottomrule
  \end{tabular}
  \caption{eMEM default configuration parameters.}
  \label{tab:config-defaults}
\end{table}

\end{document}